\documentclass[10pt,twocolumn,letterpaper]{article}

\usepackage{cvpr}
\usepackage{times}
\usepackage{epsfig}
\usepackage{graphicx}
\usepackage{amsmath}
\usepackage{amssymb}

\usepackage{times}
\usepackage{epsfig}
\usepackage{graphicx}
\usepackage{amsmath}
\usepackage{amssymb}

\usepackage{subfigure}
\usepackage{amsmath,bm}
\usepackage{array}
\usepackage{booktabs}       
\usepackage{amsfonts}       
\usepackage{nicefrac}       
\usepackage{microtype}      
\usepackage{color}      

\usepackage{multirow}

\usepackage{algorithm}
\usepackage{algorithmicx}
\usepackage{algpseudocode}
\usepackage{amsmath}

\usepackage[switch]{lineno}

\def\eg{{\em e.g.}}
\def\ie{{\em i.e.}}

\newcommand*{\affaddr}[1]{#1}
\newcommand*{\affmark}[1][*]{\textsuperscript{#1}}

\usepackage[breaklinks=true,bookmarks=false]{hyperref}

\cvprfinalcopy 

\begin{document}

\title{Guided Attention Network for Object Detection and Counting on Drones}



\author{%
Yuanqiang Cai\affmark[1,]\affmark[2], Dawei Du\affmark[3], Libo Zhang\affmark[1]\thanks{Corresponding author: Libo Zhang($\tt\small{libo@iscas.ac.cn}$). This work is supported by the National Natural Science Foundation of China under Grant No.61807033, and the Key Research Program of Frontier Sciences, CAS, Grant No.ZDBS-LY-JSC038.}, Longyin Wen\affmark[4], Weiqiang Wang\affmark[2], Yanjun Wu\affmark[1], SiweiLyu\affmark[3]\\
\affaddr{\affmark[1] Institute of Software Chinese Academy of Sciences, China}\\
\affaddr{\affmark[2] University of Chinese Academy of Sciences, China}\\
\affaddr{\affmark[3] University at Albany, SUNY, USA}\\
\affaddr{\affmark[4] JD Digits, USA}\\
}

\maketitle

\begin{abstract}
   Object detection and counting are related but challenging problems, especially for drone based scenes with small objects and cluttered background. In this paper, we propose a new Guided Attention Network (GANet) to deal with both object detection and counting tasks based on the feature pyramid. Different from the previous methods relying on unsupervised attention modules, we fuse different scales of feature maps by using the proposed weakly-supervised Background Attention (BA) between the background and objects for more semantic feature representation. Then, the Foreground Attention (FA) module is developed to consider both global and local appearance of the object to facilitate accurate localization. Moreover, the new data argumentation strategy is designed to train a robust model in various complex scenes. Extensive experiments on three challenging benchmarks (\ie, UAVDT, CARPK and PUCPR+) show the state-of-the-art detection and counting performance of the proposed method compared with existing methods.
\end{abstract}

\section{Introduction}
Object detection and counting are fundamental techniques in many applications, such as scene understanding, traffic monitoring and sports video, to name a few. However, these tasks become even more challenging in drone based scenes because of various factors such as small objects, scale variation and background clutter. With the development of deep learning, much progress has been achieved recently. Specifically, deep learning based detection and counting frameworks focus on discriminative feature representation of the objects.

First of all, the feature pyramid is widely applied in deep learning because it has rich semantics at all levels, \eg, U-Net \cite{DBLP:conf/miccai/RonnebergerFB15}, TDM \cite{DBLP:journals/corr/ShrivastavaSMG16} and FPN \cite{DBLP:conf/cvpr/LinDGHHB17}. To better exploit multi-scale feature representation, the researchers use various attention modules to fuse feature maps. In~\cite{DBLP:conf/cvpr/HuSS18}, the channel-wise feature responses are recalibrated adaptively by explicitly modelling interdependencies between channels. \cite{DBLP:conf/cvpr/0004GGH18} propose the non-local network to capture long-range dependencies, which computes the response at a position as a weighted sum of the features at all positions. Moreover, \cite{DBLP:journals/corr/abs-1904-11492} develop a lightweight global context (GC) block based on the non-local module. However, all the above methods use unsupervised attention module, but consider little about the background discriminative information in feature maps.

Based on the fused feature maps, the object is represented by proposals in anchor based methods \cite{DBLP:conf/nips/RenHGS15,DBLP:conf/eccv/LiuAESRFB16,DBLP:conf/nips/DaiLHS16} or keypoints in anchor-free methods~\cite{DBLP:conf/eccv/LawD18,DBLP:journals/corr/abs-1904-07850,DBLP:journals/corr/abs-1904-11490}. Anchor based methods exploit the global appearance information of the object, relying on pre-defined anchors. It is not flexible to design different kinds of anchors because of large scale variation in drone based scenes. Anchor-free methods employ corner points, center points or target part points to capture local object appearance without anchors. However, local appearance representation does not contain object's structure information, which is less discriminative in cluttered background, especially for small objects.

In addition, the diversity of training data is essential in deep learning. Especially in the drone based scenes, the number of difficult samples is
very limited. It is difficult for traditional data argumentation such as rescale, horizontal flip, rotation and cropping to train a robust model to deal with unconstrained drone based scenarios.

\begin{figure*}[t]
\centering
\includegraphics[width=0.95\linewidth]{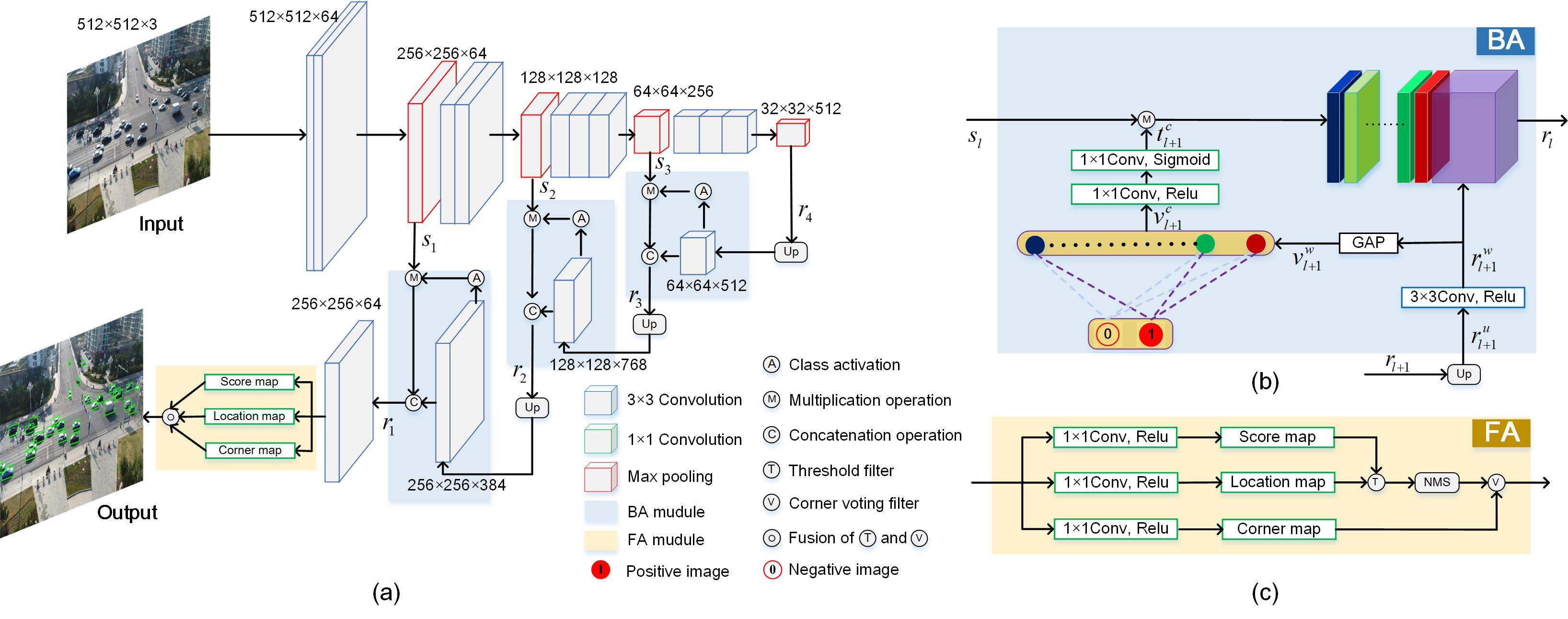}
\caption{(a) The architecture of GANet. (b) The background attention module. (c) The foreground attention module. In (a), $s_{1}$, $s_{2}$, and $s_{3}$ denote \emph{pool1}, \emph{pool2}, and \emph{pool3} low-level features, respectively; $r_{1}$, $r_{2}$, and $r_{3}$ denote the corresponding high-level features. In (b), $s_{l}$ denotes the low-level features with rich texture details, $r_{l+1}$ and $r_{l}$ denote the high-level features with strong semantic information. }
\label{fig:framework}
\end{figure*}

To address these issues, in this paper, we propose an anchor-free Guided Attention Network (GANet). First, the background attention module can enforce different channels of feature maps to learning discriminative background information for the feature pyramid. We fuse the multi-level features with the weakly-supervision of classification between background and foreground images. Second, the foreground attention module is used to capture both global and local appearance representation of the objects by tacking the merits of both anchor-based and anchor-free methods. We extract more context information in the corner regions of the object to consider local appearance information. Third, we develop a new data argumentation strategy to reduce the influence of different illumination conditions on the images for the drone based scenes, \eg, sunny, night, cloudy and foggy scenes. We conduct extensive experiments on three challenging datasets (\ie, UAVDT~\cite{DBLP:conf/eccv/DuQYYDLZHT18}, CARPK~\cite{DBLP:conf/iccv/HsiehLH17} and PUCPR+~\cite{DBLP:conf/iccv/HsiehLH17}) to show the effectiveness of the proposed method.

The main contributions of this paper are summarized as follows. (1) We present a guided attention network for object detection and counting on drones, which is formed by the foreground and background attention blocks to extract the discriminative features for accurate results. (2) A new data augmentation strategy is designed to boost up the model performance. (3) Extensive experiments on three challenging dataset, \ie, UAVDT, CARPK and PUCPR+, demonstrate the favorable performance of the proposed method against the state-of-the-arts.

\section{Guided Attention Network}
In this section, we introduce the novel anchor-free deep learning network for object detection and counting in drone images, the Guided Attention Network (GANet), which is illustrated in Figure \ref{fig:framework}. Specifically, GANet consists of three parts, \ie, the backbone, multi-scale feature fusion, and output predictor. We will first describe each part in detail, and then loss function and data argumentation strategy.

\subsection{Backbone Network}
Since diverse scales of objects are taken into consideration in feature representation, we choose the feature maps from four side-outputs of the backbone network (\eg, VGG-16~\cite{DBLP:journals/corr/SimonyanZ14a} and ResNet-50~\cite{DBLP:conf/cvpr/HeZRS16}). Four side outputs correspond to \emph{pool1}, \emph{pool2}, \emph{pool3}, and \emph{pool4}, each of which is the output of four convolution blocks with different scales, respectively. The feature maps from four \emph{pooling} layers are $\frac{1}{2}$, $\frac{1}{4}$, $\frac{1}{8}$, $\frac{1}{16}$ the size of the input image. They are marked with light blue regions in Figure~\ref{fig:framework}(a). The backbone network is pre-trained by the ImageNet dataset~\cite{DBLP:conf/nips/KrizhevskySH12}.

\subsection{Multi-Scale Feature Fusion}
As discussed in~\cite{DBLP:conf/cvpr/LinDGHHB17}, the feature pyramid has strong semantics at all scales, resulting in significant improvement as a generic feature extractor. Specifically, we fuse the side-outputs of the backbone network from top to down, \eg, feature maps from \emph{pool4} to \emph{pool1} of VGG-16. Meanwhile, the receptive fields of the stacked feature maps can adaptively match the scale of objects. To consider background discriminative information in the feature pyramid, we introduce the Background Attention (BA) module in multi-scale feature fusion.

\subsubsection{Background Attention.}
As shown in Figure~\ref{fig:framework}(b), the BA modules are stacked from the deepest to the shallowest convolutional layer. At the same time, the cross-entropy loss function is used to enforce different channels of feature maps focus on either foreground and background in every stage. Then, the attention module weights the pooling features with the same scale via the class-activated tensor. Finally, the weighted pooling features and the up-sampled features are concatenated and regarded as the base feature maps in the next BA.

\begin{figure}[t]
\centering
\includegraphics[width=\linewidth]{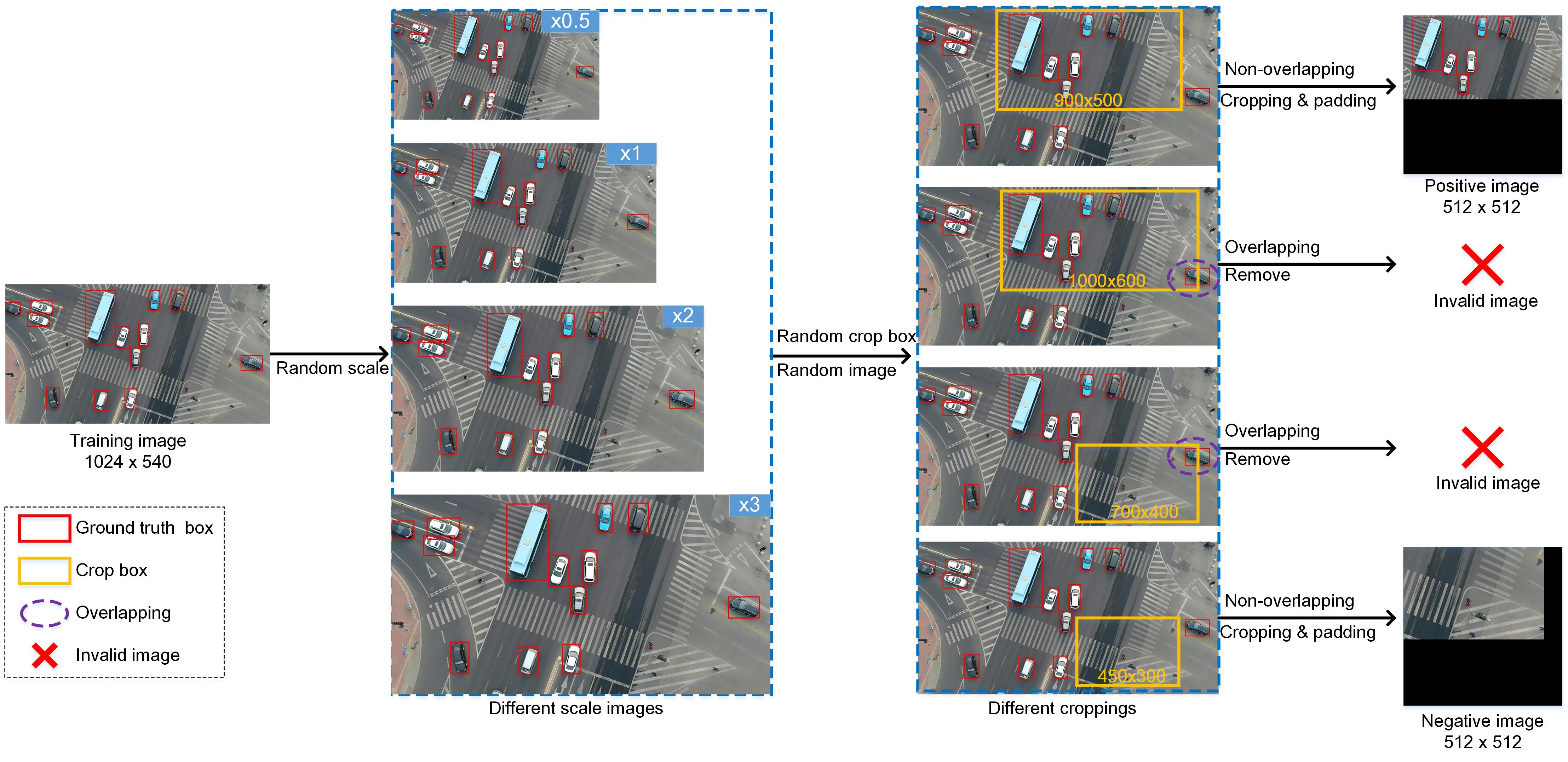}
\caption{Generation of positive and negative samples.}
\label{fig:generation}
\end{figure}

We denote the $l$-th pooling features as $s_l$, and the input and output of $l$-th BA as $r_{l+1}$ and $r_{l}$. Specifically, $r_{l+1}$ is used to learn the class-related weights for activating the class-related feature maps in $s_{l}$. For the deepest BA module, the input is regarded as the \emph{pool4} feature maps (see \emph{$r_{4}$} in Figure~\ref{fig:framework}(a)). Note that the size of output $r_{l}$ in this architecture is the same as the pooling features $s_{l}$ rather than the size of input $r_{l+1}$. Therefore, the bilinear interpolation is introduced to up-sample $r_{l+1}$ to $r^{u}_{l+1}$. As the up-sampling operation is a linear transformation, one $3\times3$ convolutional layer $w^{u}_{l}$ is used as soft-adding to improve the scale adaptability. Instead of concatenating the up-sampled $r_{l+1}$ and the activated $s_{l}$ directly, the $1\times1$ and $3\times3$ convolutional layers $w^{c}_{l}$ is used to generate $r_{l}$. In summary, the $l$-th BA is formulated as
\begin{equation}
\label{equ:SAU1}
r_{l} = w^{c}_{l}\cdot(f(s_{l},r^{w}_{l+1})+r^{w}_{l+1}),
\end{equation}
where $w^{c}_{l}$ denotes the convolutional weights of the concatenation layer. $r^{w}_{l+1}=w^{u}_{l}\ast r^{u}_{l+1}$ and $w^{u}_{l}$ are the convolutional weights of up-sampled $r^{u}_{l+1}$. $w^{c}_{l}$ has two elements, \ie, one for $r^w_{l+1}$ and the other for $f(s_l,r^w_{l+1})$. $f(s_{l},r^{w}_{l+1})$ is a class activation function with two parameters, \ie, the pooling features $s_{l}$ and the weighted up-sampled features $r^{w}_{l+1}$. It is defined as
\begin{equation}
\label{equ:SAU2}
f(s_{l},r^{w}_{l+1}) = s_{l} \otimes g^{c}(r^{w}_{l+1}),
\end{equation}
where $\otimes$ is the multiply operation between the features $s_{l}$ and the weight tensor $g^{c}(r^{w}_{l+1})$. $g^{c}(r^{w}_{l+1})$ is obtained by three steps. First, $r^{w}_{l+1}$ is compressed into a one-dimensional vector $v^{w}_{l+1}$ by the Global Average Pooling (GAP)~\cite{DBLP:conf/cvpr/ZhouKAHE16}. Second, $v^{w}_{l+1}$ is activated and converted to the vector with class-related information $v^{c}_{l+1}$ via determining whether the input image contains the objects. Third, $v^{c}_{l+1}$ is transformed into a weight tensor with class-related information $t^{c}_{l+1} = g^{c}(r^{w}_{l+1})$ via two $1\times1$ convolutional layers.

\subsubsection{Positive and Negative Image Generation.}
To learn class-related feature maps, we use both the images with and without objects in the training stage. We denote them as positive and negative images respectively. Specifically, we use positive images with objects to activate the channels of feature maps to represent the pixels of object region, and negative images without overlapping of objects to activate the channels of feature maps to describe the background region. As shown in Figure~\ref{fig:generation}, we generate positive and negative images with the size of $512\times512$ by randomly cropping and padding the rescaled training images (from $0.5$x to $3$x scale).

\subsection{Output Predictor}
Based on multi-scale feature fusion, we predict the scales and locations of objects using both score and location maps (see Figure~\ref{fig:framework}(c)), which are defined as follows:
\begin{itemize}
\item  The score map corresponds to confidence score of the object region. Similar to the confidence map in FCN~\cite{DBLP:conf/cvpr/LongSD15}, each pixel of the score map is a scalar between $0$ to $1$ representing the confidence belonging to an object region.
\item  The location map describes the location of object by using four distance channels $G=(l,t,r,b)$. The channels denote the distances from the current pixel $i$ to the \emph{left}, \emph{top}, \emph{right}, and \emph{bottom} edges of the bounding box respectively. Then we can directly predict the object box by four distance channels. Specifically, for each point in the score map, four distance channels predict the distances to the above four edges of the bounding box.
\end{itemize}

\subsubsection{Foreground Attention.}
In general, based on both score and location maps, we can estimate the bounding boxes of the objects in the image. However, the estimated bounding boxes only rely on the global appearance of the object. That is, little local appearance of the object is taken into consideration, resulting in less discriminative foreground representation. To improve localization accuracy, we introduce the Foreground Attention (FA) module to consider both global and local appearance representation of the objects.

\begin{figure}[t]
\centering
\includegraphics[width=\linewidth]{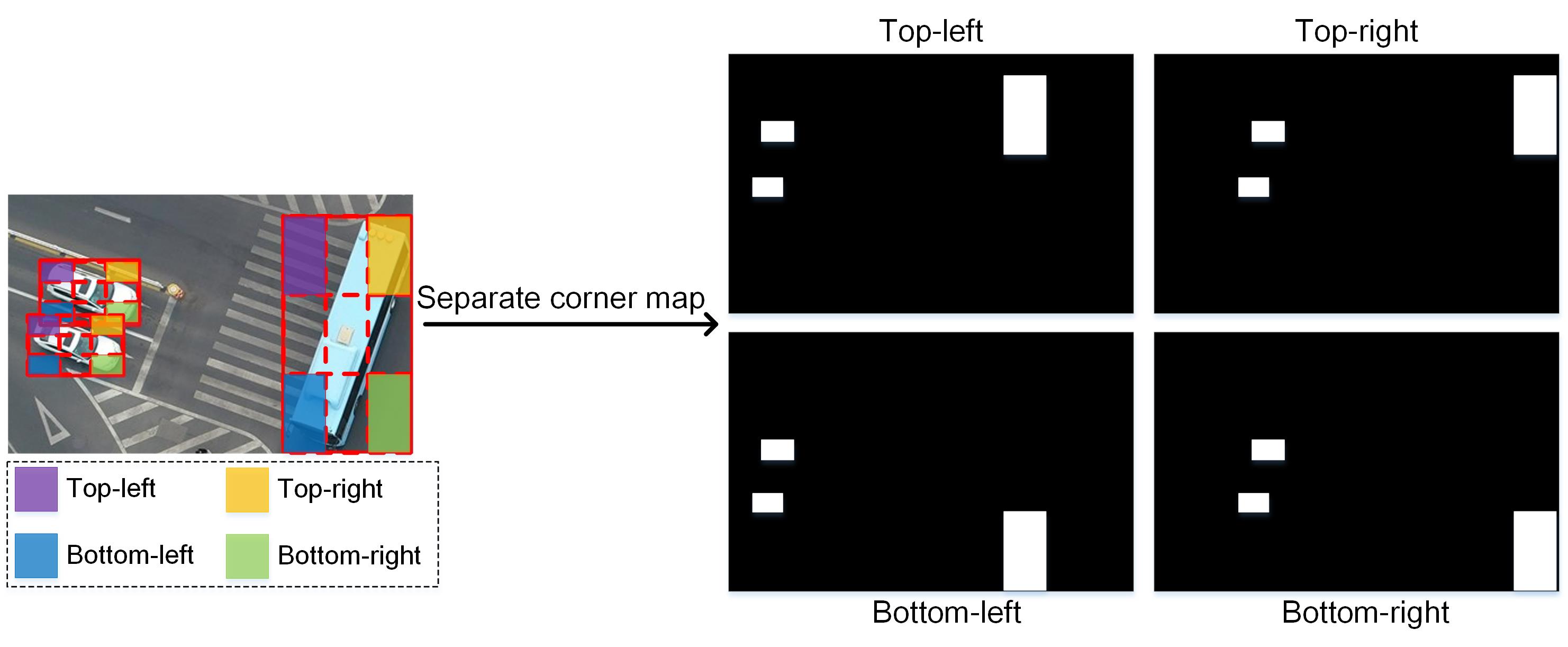}
\caption{Illustration of four corner maps for foreground attention.}
\label{fig:corner}
\end{figure}

In practice, we use four corner maps (\textit{top-left}, \textit{top-right}, \textit{bottom-left} and \textit{bottom-right}) to denote different corner positions within the object region, as shown in Figure~\ref{fig:corner}. Similar to score map, each pixel of the corner map is also a scalar between $0$ to $1$ representing the confidence belonging to a corresponding position in the object region. The corner is set as $1/9$ the size of the whole object. Specifically, as illustrated in Figure~\ref{fig:framework}(c), we first use a threshold filter to remove the candidate bounding boxes with low confidence pixels, \ie, $c_{i}<\mu$. $c_{i}$ is the confidence value of pixel $i$ in the predicted score map, and $\mu$ denotes the confidence threshold. Then, the Non-Maximum Suppression (NMS) operation is applied to remove redundant candidate bounding boxes and choose the top ones with higher confidence. Finally, a corner voting filter is designed to determine whether the selected bounding boxes should be retained. Specifically, we calculate the number of reliable corners $\mathbb{N}(b_{k})$ in the $k$-th candidate bounding box $b_{k}$ by
\begin{equation}
\label{equ:corner}
\mathbb{N}(b_{k}) = \sum_{s=1}^{4}\mathbb{I}(\tau(\mathcal{C}_{s})>\varepsilon),
\end{equation}
where $\tau(\mathcal{C}_{s})$ denotes the average confidence of the corner region $\mathcal{C}_{s}$. $\varepsilon$ indicates the threshold of mean confidence $\tau(\mathcal{C}_{s})$ to determine the reliable corner. $\mathbb{I}(\cdot) = 1$ if its argument is true, and $0$ otherwise. We only keep the bounding box $b_{k}$ if the number of reliable corners is larger than the threshold $\kappa$, \ie, $\mathbb{N}(b_{k})>\kappa$.

\subsection{Loss function}
To train the proposed network, We optimize the location map and score map, as well as both foreground and background attentions simultaneously. The overall loss function is defined as
\begin{equation}
\mathcal{L} = \mathcal{L}_\text{loc} + \lambda_\text{sco}\mathcal{L}_\text{sco} + \lambda_\text{FA}\mathcal{L}_\text{FA} + \lambda_\text{BA}\mathcal{L}_\text{BA},
\label{eq:loss}
\end{equation}
where $\mathcal{L}_\text{loc}$, $\mathcal{L}_\text{sco}$, $\mathcal{L}_\text{FA}$, and $\mathcal{L}_\text{BA}$ are loss terms for the location map, score map, foreground attention, and background attention, respectively. The parameter $\lambda_\text{sco}$, $\lambda_\text{FA}$, and $\lambda_\text{BA}$ are used to balance these terms. In the following, we explain these loss terms in detail.

\subsubsection{Loss of Location Map.}
To achieve scale-invariance, the IoU loss~\cite{DBLP:conf/mm/YuJWCH16} is adopted to evaluate the difference between the predicted bounding box and the ground truth of bounding box. The loss of location map is defined as:
\begin{equation}
\label{loss:HRN_Geometry}
\small
\mathcal{L}_\text{loc}=\text{IoU}(G,G^{\ast}),
\end{equation}
where $G=(l,t,r,b)$ and $G^{\ast}=(l^{\ast},t^{\ast},r^{\ast},b^{\ast})$ are the estimated and ground-truth bounding box of the object. The function $\text{IoU}(\cdot)$ calculates the intersection-over-union (IoU) score between $G$ and $G^{\ast}$.

\subsubsection{Loss of Score Map.}
Similar to image segmentation~\cite{DBLP:journals/corr/abs-1712-09093}, we use the Dice loss to deal with the imbalance problem of positive and negative pixels in the score map. It calculates the errors between the predicated score map and ground-truth map. The loss is calculated as
\begin{equation}
\small
\mathcal{L}_\text{sco} = 1- \frac{2\cdot\sum_{i=1}^{N}(c_{i}c^{\ast}_{i})}{\sum_{i=1}^{N}(c_{i})+\sum_{i=1}^{N}(c^{*}_{i})},
\end{equation}
where the sums run over the all $N$ pixels of the score map. $c^{\ast}_{i}$ and $c_{i}$ are the confidence values of pixel $i$ in the ground-truth and predicted maps respectively.

\begin{figure}[t]
  \centering
  \subfigure[]{
  \centering
    \includegraphics[width=\linewidth]{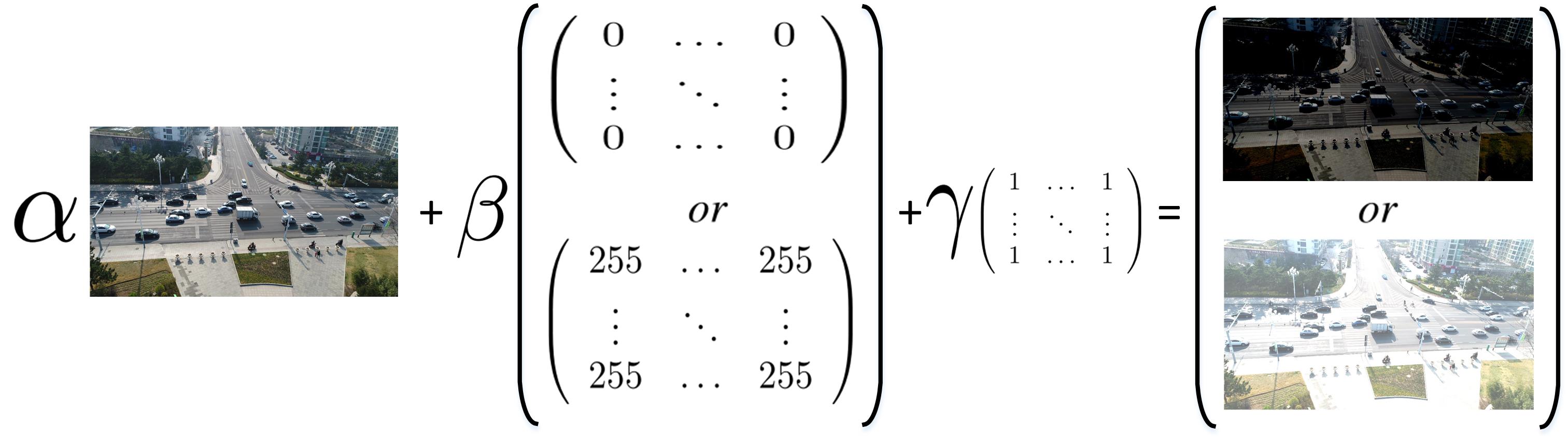}
  \label{fig:DA_BNoise}
  }
  \subfigure[]{
  \centering
    \includegraphics[width=\linewidth]{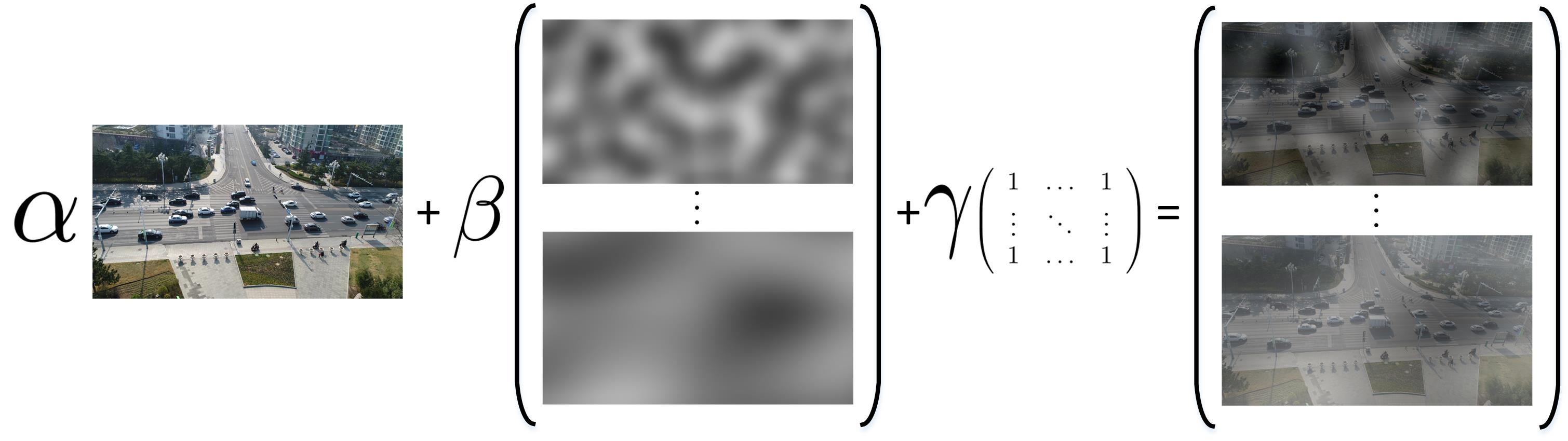}
  \label{fig:DA_PNoise}
  }
\centering
  \caption{Illustration of data augmentation including (a) BNoise (Brightness noises to imitate sunny or night scenes) and (b) PNoise (Perlin noises to imitate cloudy and foggy scenes).}
\label{fig:DataAugmentation}
\end{figure}

\subsubsection{Loss of Background Attention.}
Similar to classification algorithms, we use the cross-entropy loss $\mathcal{L}_\text{BA}$ to guide background attention based on the binary classification, \ie,
\begin{align}
\small
\mathcal{L}_\text{BA} = \left\{ \begin{array}{ll}
-\log(p) & \text{if}\quad y=1,\\
-\log(1-p) & \text{otherwise},
\end{array} \right.
\end{align}
where $y\in\{\pm1\}$ denotes the ground-truth category (\ie, foreground or background), $p\in[0,1]$ is the estimated probability for the category with label $y=1$.

\subsubsection{Loss of Foreground Attention.}
Similar to the score map, to deal with the imbalance problem of positive and negative pixels in the feature maps, we use the Dice loss to guide the foreground attention for the four corner maps.

\subsection{Data Augmentation for Drones}
Data augmentation is important in deep network training based on limited training data. Since the data is captured from a very high altitude by the drone, it is susceptible to the influence of different illumination conditions, \eg, sunny, night, cloudy and foggy. Therefore, we develop a new data augmentation strategy for drones.

As we know, sunny or night scenes correspond to the brightness of the image, therefore we synthesize these scenes via changing the whole contrast of the image (denoted as \textit{BNoise}). On the other hand, since convincing representations of clouds and water can be created in pixel-level~\cite{DBLP:conf/siggraph/Perlin85}, we use Perlin noise~\cite{DBLP:journals/tog/Perlin02} to imitate cloudy and foggy scenes (denoted as \textit{PNoise}). Inspired by the image blending algorithm~\cite{szeliski2010computer}, the data augmentation model is defined as
\begin{equation}
\Phi(i) = \alpha I(i) + \beta M^{\ast}(i) + \gamma,
\label{eq:aug}
\end{equation}
where $\Phi(i)$ is the transformed value of the pixel $i$ in image. $\alpha$ and $\beta$ denote the weight of the pixel of original image $I(p)$ and noise map $M^{\ast}(i)$ respectively. The asterisk $\ast$ denotes different kinds of noise maps, \ie, BNoise $M^b(i)$ and PNoise $M^p(i)$. We have $\alpha = 1 - \beta$ to control the contrast of the image. The perturbation factor $\gamma$ is used to revise the brightness. We set different factors $\alpha$ and $\gamma$ for each image in the training phase.

As shown in Figure~\ref{fig:DA_BNoise}, we employ white and black maps to synthesize sunny or night images. On the other hand, we use Perlin noise~\cite{DBLP:journals/tog/Perlin02} to generate noise maps in Figure~\ref{fig:DA_PNoise}, and then revise the brightness via disturbance factor $\gamma$ to synthesize cloudy and foggy images. For each training image, we first resize it using random scale factors (x$0.5$, x$1$, x$2$ and x$3$). Then, we introduce both noise maps into the image to imitate the challenging scenes (\ie, sunny, night, cloudy, and foggy). Finally, we select positive and negative images by random cropping on the blending images, and transform the selected images to $512\times512$ size via zooming and padding.

\section{Experiment}
\begin{figure*}[t]
  \centering
  \subfigure[UAVDT]{
  \centering
    \includegraphics[width=3.2in,height=2.1in]{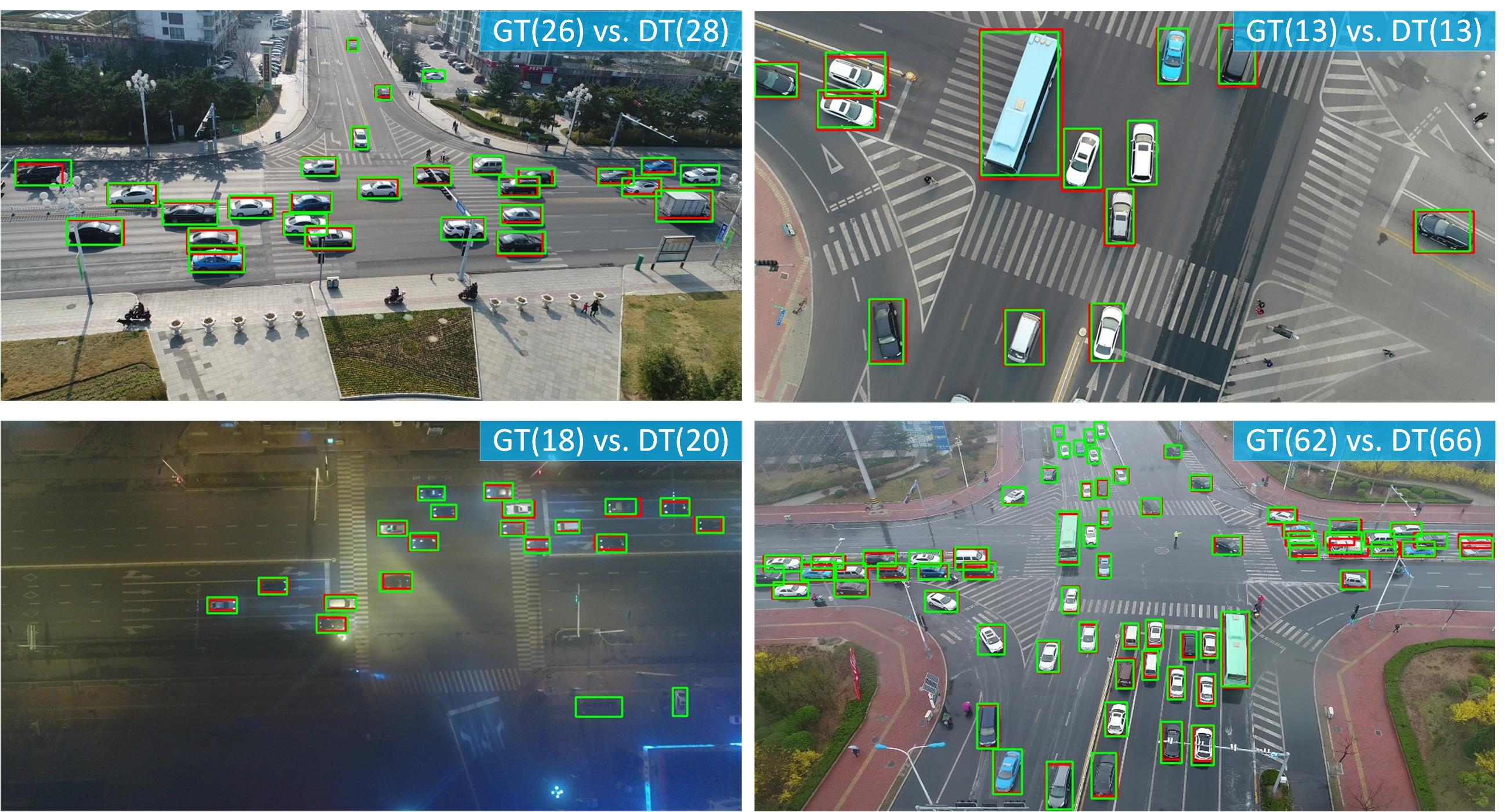}
  \label{fig:SAN_Examples11_UAVDT}
  }
  \subfigure[CARPK]{
  \centering
    \includegraphics[width=1.6in,height=2.1in]{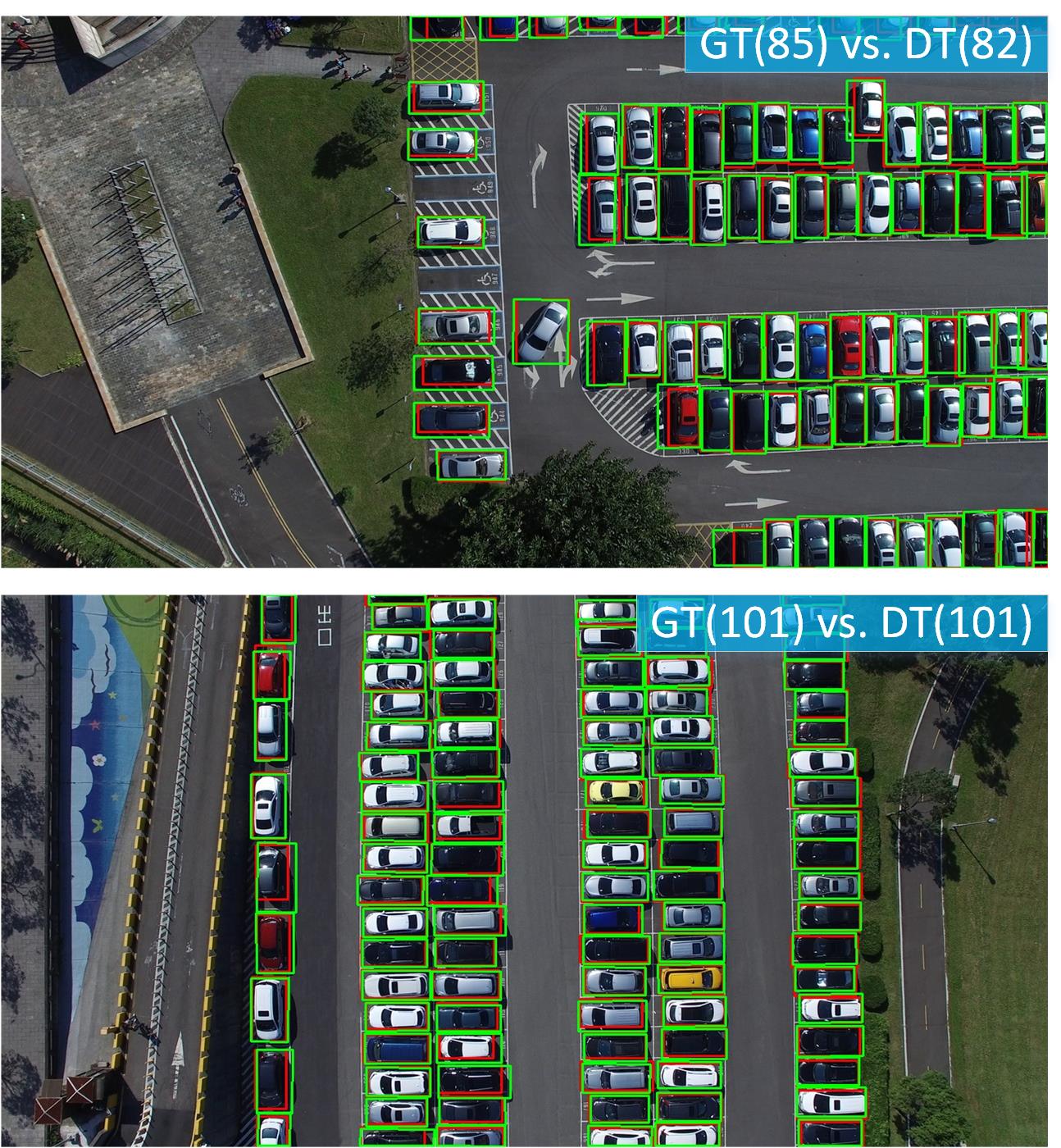}
  \label{fig:SAN_Examples11_CARPK}
  }
  \subfigure[PUCPR+]{
  \centering
    \includegraphics[width=1.6in,height=2.1in]{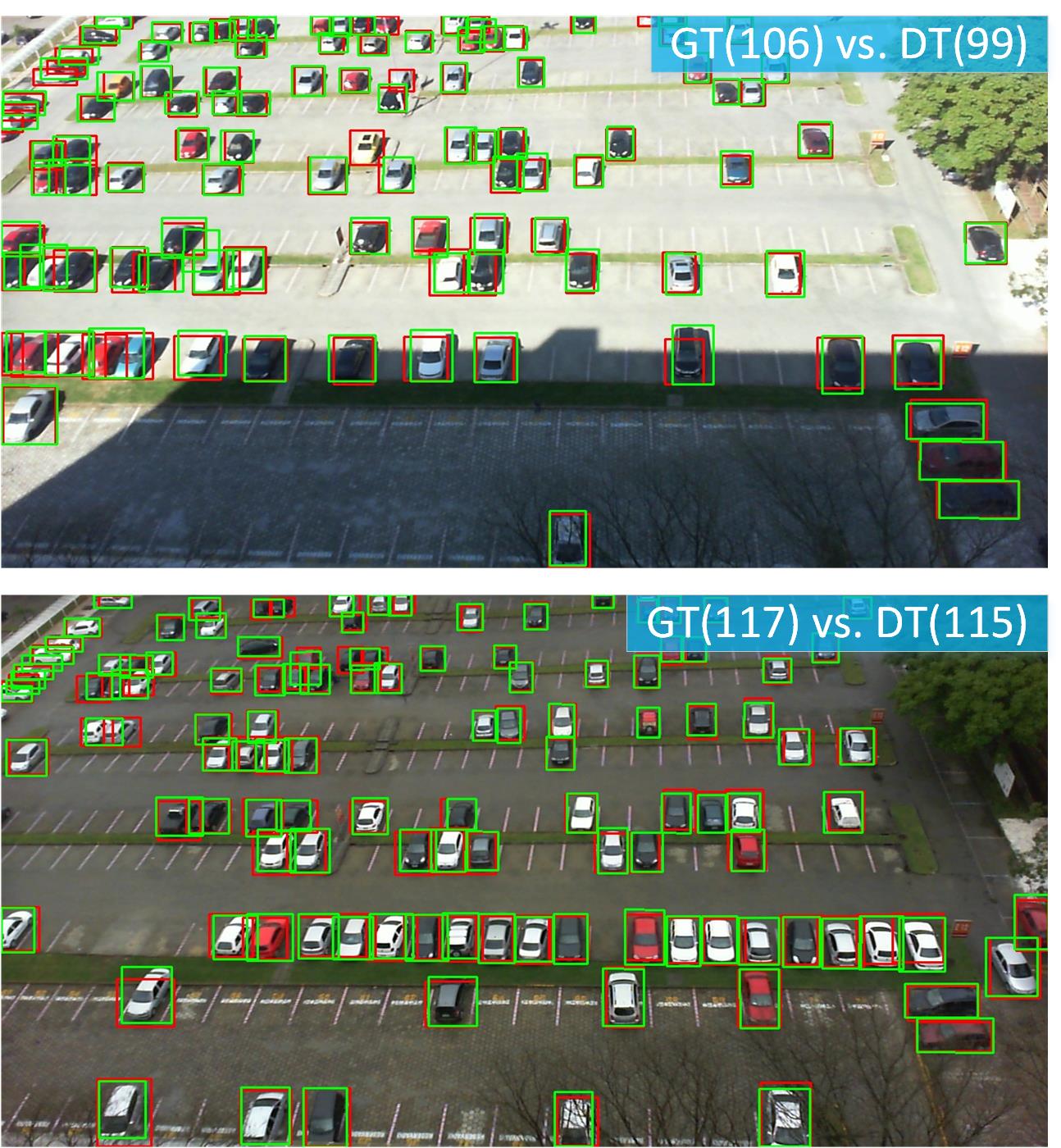}
  \label{fig:SAN_Examples11_PUCPRPlus}
  }
  \centering
  \caption{Visual examples of GANet with VGG16 backbone. The ground-truth and predicted detection bounding boxes are highlighted in red and green rectangles, respectively. The blue mask in the top-right corner indicates the comparison between the ground-truth (GT) and estimated detection (DT) counts.}
\label{fig:example}
\end{figure*}

The proposed method is implemented by Tensorflow r1.8\footnote{\url{https://www.tensorflow.org/}}. We will release the source codes of our method upon the acceptance of the paper. We evaluate our method on two drone based datasets: UAVDT~\cite{DBLP:conf/eccv/DuQYYDLZHT18} and CARPK~\cite{DBLP:conf/iccv/HsiehLH17}. We also evaluate our method on the PUCPR+ dataset~\cite{DBLP:conf/iccv/HsiehLH17} because the dataset is collected from the $10$th floor of a building and similar to drone view images to a certain degree. In this section, we first describe implementation details. Then, we compare our GANet with the state-of-the-art methods, \ie, Faster R-CNN \cite{DBLP:conf/nips/RenHGS15}, RON~\cite{DBLP:conf/cvpr/KongSYLLC17}, SSD~\cite{DBLP:conf/eccv/LiuAESRFB16}, R-FCN~\cite{DBLP:conf/nips/DaiLHS16}, CADNet~\cite{duan2019detecting}, One-Look Regression~\cite{DBLP:conf/eccv/MundhenkKSB16}, IEP~\cite{DBLP:journals/tip/StahlPG19}, YOLO9000~\cite{DBLP:conf/cvpr/RedmonF17}, LPN~\cite{DBLP:conf/iccv/HsiehLH17}, RetinaNet~\cite{DBLP:conf/iccv/LinGGHD17}, YOLOv3~\cite{DBLP:journals/corr/abs-1804-02767}, IoUNet~\cite{DBLP:journals/corr/abs-1904-00853}, and SA+CF+CRT~\cite{DBLP:journals/tie/LiLWCN19}. More visual examples are shown in Figure~\ref{fig:example}. In addition, the ablation study is carried out to evaluate the effectiveness of each component in our network.

\subsubsection{Implementation Details}
Due to the shortage of computational resources, we train GANet using the VGG-16 and ResNet-50 backbone with the input size $512\times512$. All the experiments are carried out on the machine with NVIDIA Titan Xp GPU and Intel(R) Xeon(R) E5-1603v4@2.80GHz CPU. For fair evaluation, we generate the same top $200$ detection bounding boxes for the UAVDT and CARPK datasets and $400$ detection bounding boxes for the PUCPR+ dataset based on the detection confidence. Note that the detection confidence is calculated by summarizing the value of each pixel in the score map. To output the count of objects in each image, we calculate the number of detection with the detection confidence larger than $0.5$. We fine-tune the resulting model using the Adam Optimizer. An exponential decay learning rate is used in the training phrase, \ie, its initial value is $0.0001$ and decays every $10,000$ iterations with the decay rate $0.94$. The batch size is set as $10$. In the loss function~\eqref{eq:loss}, we set the balancing factors as $\lambda_{sco}=0.01$, $\lambda_\text{FA}=0.0025$, $\lambda_\text{BA}=0.001$ empirically. In the FA module, the confidence threshold $\mu$ is set as $0.8$, and the threshold $\varepsilon$ in \eqref{equ:corner} is set as $0.3$ empirically. The Non-Maximum Suppression (NMS) operation is conducted with a threshold $0.2$. In the data argumentation model~\eqref{eq:aug}, we set the balancing weights as $\alpha=\{0.1,0.3,0.5,0.7,0.8,0.9,1.0\}$ and $\gamma=[-20,20]$.

\subsubsection{Metrics.}
To evaluate detection algorithms on the UAVDT dataset~\cite{DBLP:conf/eccv/DuQYYDLZHT18}, we compute the Average Precision (AP$@0.7$) score based on~\cite{DBLP:conf/cvpr/GeigerLU12,DBLP:journals/ijcv/EveringhamEGWWZ15}. That is, the hit/miss threshold of the overlap between detection and ground-truth bounding boxes is set to $0.7$. In terms of CARPK~\cite{DBLP:conf/iccv/HsiehLH17} and PUCPR+~\cite{DBLP:conf/iccv/HsiehLH17}, we report the detection score under two hit/miss thresholds, \ie, AP$@0.5$ and AP$@0.7$. To evaluate the counting results, similar to~\cite{DBLP:conf/iccv/HsiehLH17}, we use two object counting metrics including Mean Absolute Error (MAE) and Root Mean Squared Error (RMSE).

\subsection{Quantitative Evaluation}
\begin{table}[t]
\small
\begin{center}
\caption{Comparison on the UAVDT dataset.}
\label{tab:res_UAVDT}
\setlength{\tabcolsep}{4.0pt}
\begin{tabular}{l c cc c}
\hline
\toprule
Method        & Backbone   & MAE$\downarrow$ & RMSE$\downarrow$  & AP$@0.7$[$\%$]$\uparrow$   \\
\midrule
YOLO9000    &DarkNet-19 &12.59  &16.73   &7.6 \\
YOLOv3      &DarkNet-53 &11.58  &21.50   &20.3  \\
RON    & VGG-16     &- &-   &21.6  \\
Faster R-CNN        &VGG-16 &-&-    &22.3   \\
SSD &VGG-16   &-&-   &33.6     \\
CADNet   &VGG-16   &-&-   &43.6     \\
Ours & VGG-16  &\textbf{5.10}  & \textbf{8.10}   &\textbf{46.8} \\
\midrule
SA+CF+CRT  &ResNet-101  &7.67   &10.95  &27.8 \\
R-FCN   &ResNet-50   &-&-  &34.4 \\
Ours &ResNet-50  &\textbf{5.09}&\textbf{8.16} &\textbf{47.2}\\
\bottomrule
\end{tabular}
\end{center}
\end{table}

\begin{table}[t]
\begin{center}
\caption{Comparison on the CARPK dataset.}
\label{tab:res_CARPK}
\setlength{\tabcolsep}{0.5pt}
\footnotesize{
\begin{tabular}{lcccc}
\hline
\toprule
Method & MAE$\downarrow$ & RMSE$\downarrow$ & AP@0.5[$\%$]$\uparrow$ &AP@0.7[$\%$]$\uparrow$\\
\midrule
One-Look Regression &59.46 &66.84&-&- \\
IEP &51.83 &-&-&- \\
Faster R-CNN      &47.45      &57.39 &-&-   \\
YOLO9000    &38.59      &43.18 &20.9 &3.7   \\
SSD    &37.33      &42.32 &68.7 &25.9   \\
LPN&23.80    &36.79    &-&-\\
RetinaNet &16.62       &22.30 &- &-     \\
YOLOv3&7.92      &11.08 &85.3 &47.0  \\
IoUNet &6.77      &8.52&-&-     \\
SA+CF+CRT &5.42    &7.38 &89.8 &61.4  \\
\midrule
Ours (VGG-16) &4.80    &6.94   &\textbf{90.2} &73.6 \\
Ours (ResNet-50) &\textbf{4.61}    &\textbf{6.55} &90.1&\textbf{74.9}   \\
\bottomrule
\end{tabular}}
\end{center}
\end{table}

\subsubsection{Evaluation on UAVDT.}
The UAVDT dataset~\cite{DBLP:conf/eccv/DuQYYDLZHT18} consists of $100$ video sequences with approximate $80,000$ frames, which are collected from various scenes. Moreover, the objects are annotated by bounding boxes as well as several attributes (\eg, weather condition, flying altitude, and camera view). Note that we only use the subset of UAVDT dataset for object detection in our experiment. As presented in Table~\ref{tab:res_UAVDT}, we can conclude that our GANet performs the best among all the compared detection methods in terms of both the VGG-16 and ResNet-50 backbones. Specifically, GANet surpasses YOLO9000, YOLOv3, RON, Faster R-CNN, SSD, CADNet, SA+CF+CRT and R-FCN by $39.2\%$, $26.3\%$ $25.2\%$, $24.5\%$, $13.2\%$, $3.2\%$, $19.4\%$ and $12.8\%$ AP scores, respectively. Moreover, our method achieves better counting accuracy than SA+CF+CRT with the more complex ResNet-101 backbone, \ie, $5.09$ MAE score and $8.16$ RMSE score. It demonstrates that the effectiveness of our method in object detection in drone based scenes.

\begin{table}[t]
\begin{center}
\caption{Comparison on the PUCPR+ dataset.}
\label{tab:res_PUCPR}
\setlength{\tabcolsep}{0.5pt}
\footnotesize{
\begin{tabular}{lcccc}
\hline
\toprule
Method & MAE$\downarrow$ & RMSE$\downarrow$ & AP@0.5[$\%$]$\uparrow$ &AP@0.7[$\%$]$\uparrow$\\
\midrule
SSD       &119.24      &132.22   & 32.6 &7.1 \\
Faster R-CNN       &111.40      &149.35 &-&-    \\
YOLO9000   &97.96      &133.25 &12.3  &4.5  \\
RetinaNet  &24.58       &33.12  &-&-  \\
LPN &23.80    &36.79   &-&- \\
One-Look Regression &21.88 &36.73 &-&-\\
IEP&15.17 &- &-&-\\
IoUNet  &7.16      &12.00  &-&-  \\
YOLOv3&5.24      &7.14 &\textbf{95.0} &45.4    \\
SA+CF+CRT&3.92    &5.06  &92.9 &55.4  \\
\midrule
Ours (VGG-16) &3.68    &5.47  &91.3 &\textbf{67.0}  \\
Ours (ResNet-50) &\textbf{3.28}    &\textbf{4.96}  & 91.4 &65.5\\
\bottomrule
\end{tabular}}
\end{center}
\end{table}

\begin{table*}[t]
\centering
{
\caption{Comparison of variants of GANet on the UAVDT dataset.}
\label{tab:variant}
\begin{tabular}{l c ccc ccc ccc}
\hline
\toprule
Method & AP & AP$_\text{day}$ & AP$_\text{night}$ & AP$_\text{fog}$ & AP$_\text{low}$ & AP$_\text{med}$ & AP$_\text{high}$  & AP$_\text{front}$ & AP$_\text{side}$ & AP$_\text{bird}$   \\
\midrule
GANet         &0.3908      &0.4779     &0.5513  &0.1509 &0.5505&0.4616&0.1227&0.4478&0.5111&0.1981\\
GANet+BPNoise  &0.4181      &0.4940 &0.5581 & 0.2027&0.5565&0.4867&0.1665&0.4618&0.5219&0.2533\\
GANet+FA         &0.4207      &0.5006     &\textbf{0.5878}  &0.1890&\textbf{0.5935}&0.4834&0.1431&0.4595&0.5462&0.2439\\
GANet+BA   &0.4353   &0.5041 &0.5743 &0.2401&0.5908&0.4812&0.1996&0.4655&0.5451&0.2947\\
GANet+BPNoise+FA         &0.4411      &\textbf{0.5272}     &0.5819  &0.2139 &0.5900&0.5146&0.1751&0.4805&\textbf{0.5618}&0.2715\\
GANet+BPNoise+BA         &0.4576      &0.5049     &0.5779  &0.3068&0.5815&0.4923&\textbf{0.2695}&0.4719&0.5309&\textbf{0.3640}\\
GANet+BPNoise+FA+BA    &\textbf{0.4679}   &0.5240  &0.5841  &\textbf{0.3084}&0.5820&\textbf{0.5206}&0.2624&\textbf{0.4852}&0.5435&0.3603\\
\bottomrule
\end{tabular}}
\end{table*}

\begin{figure*}[t]
\centering
\includegraphics[width=0.9\linewidth]{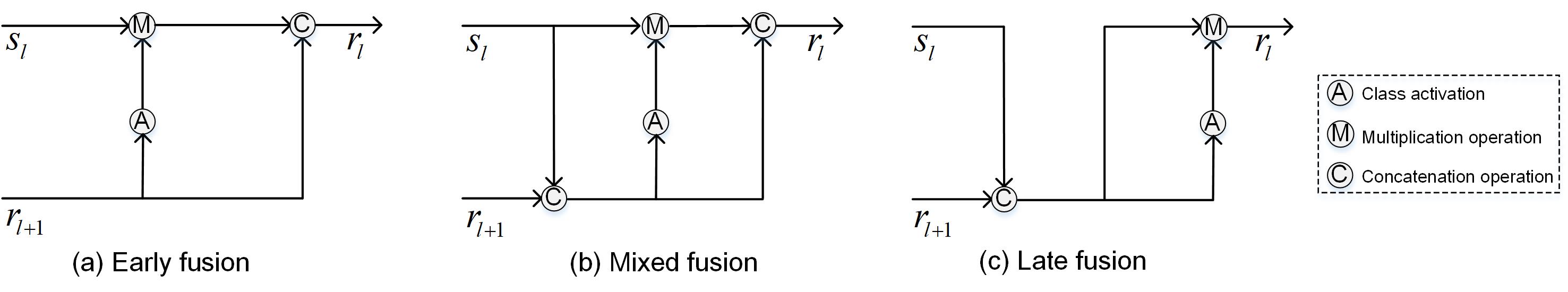}
\caption{Different fusion strategies of multi-scale feature maps. $s_{l}$ denotes the low-level features with rich texture details, $r_{l+1}$ and $r_{l}$ denote the high-level features with strong semantic information.}
\label{fig:fusion}
\end{figure*}

\subsubsection{Evaluation on CARPK.}
The CARPK dataset~\cite{DBLP:conf/iccv/HsiehLH17} provides the largest-scale drone view parking lot dataset in unconstrained scenes, which is collected in various scenes for $4$ different parking lots. It contains approximately $90,000$ cars in total with the view of drone. We compare our method with state-of-the-art algorithms in Table~\ref{tab:res_CARPK}. The results show that our approach achieves the best MAE, RMSE and AP scores. It is worth mentioning that we obtain much better AP$@0.7$ score (\ie, $74.9$ vs. $61.4$). This is attributed to the proposed attention modules to locate the objects more accurately.

\begin{table}[t]
\begin{center}
\caption{Influence of data augmentation.}
\label{tab:abl_aug}
\begin{tabular}{l cccc}
\hline
\toprule
Method & AP & AP$_\text{day}$ & AP$_\text{night}$ & AP$_\text{fog}$   \\
\midrule
GANet         &0.3908      &0.4779     &0.5513  &0.1509 \\
GANet+BNoise         &0.4034      &0.4928     &\textbf{0.5686}  & 0.1579\\
GANet+PNoise         &0.4063      &0.4798     &0.5263  & \textbf{0.2118}\\
GANet+BPNoise   &\textbf{0.4181} &\textbf{0.4940} &0.5581 & 0.2027\\
\bottomrule
\end{tabular}
\end{center}
\end{table}
\subsubsection{Evaluation on PUCPR+.}
The PUCPR+ dataset~\cite{DBLP:conf/iccv/HsiehLH17} is the subset of PKLot~\cite{DBLP:journals/eswa/AlmeidaOBJK15}, which is annotated with nearly $17,000$ cars in total. It shares the similar high altitude attribute to drone based scenes, but the camera sensors are fixed and set in the same place. As presented in Table~\ref{tab:res_PUCPR}, our method performs the best in terms of MAE and RMSE scores. YOLOv3~\cite{DBLP:journals/corr/abs-1804-02767} achieves the best AP score at $0.5$ hit/miss threshold, but inferior AP$@0.7$ score than that of our method. We speculate that YOLOv3 lack of global appearance representation of objects to achieve accurate localization.

\begin{table}[t]
\centering
\setlength{\tabcolsep}{4pt}
{
\caption{Influence of background attention.}
\label{tab:abl_BA}
\begin{tabular}{lcccc}
\hline
\toprule
Method & AP & AP$_\text{front}$ & AP$_\text{side}$ & AP$_\text{bird}$   \\
\midrule
GANet+BPNoise         &0.4181 &0.4618&0.5219&0.2533 \\
GANet+BPNoise+LF      &0.4457 &0.4667 &0.5301 & 0.3294 \\
GANet+BPNoise+MF      &0.4530 &0.4699 &\textbf{0.5338} & 0.3495 \\
GANet+BPNoise+EF      &\textbf{0.4576} &\textbf{0.4719}&0.5309&\textbf{0.3640} \\
\midrule
GANet+BPNoise+FPN              &0.3985 &0.4378 &0.4943 & 0.2480 \\
GANet+BPNoise+GC        &0.4343 &0.4681 &\textbf{0.5374} & 0.2919 \\
GANet+BPNoise+SE        &0.4442 &\textbf{0.4723} &0.5347 &0.3142 \\
GANet+BPNoise+BA       &\textbf{0.4576}  &0.4719&0.5309&\textbf{0.3640} \\
\bottomrule
\end{tabular}}
\end{table}

\subsection{Ablation Study}
We perform analyses on the effect of the important modules in our method on the detection performance. Specifically, we study the influence of data augmentation, semantic discriminative attention, and corner attention. We select the UAVDT dataset~\cite{DBLP:conf/eccv/DuQYYDLZHT18} to conduct the experiment because it provides various attributes in terms of altitude, illumination and camera-view for comprehensive evaluation.

\subsubsection{Effectiveness of Data Augmentation.}
As discussed above, the data augmentation strategy is used to increase the difficult samples affected by various illumination attributes in the UAVDT dataset~\cite{DBLP:conf/eccv/DuQYYDLZHT18} such as \textit{daylight}, \textit{night} and \textit{fog}. We compare different variants of GANet with different data augmentation, denoted as GANet+BNoise, GANet+PNoise and GANet+PBNoise. Notably, BNoise denotes the brightness noise, PNoise denotes the Perlin noise, and BPNoise denotes both. As shown in Table~\ref{tab:abl_aug}, the performance of GANet+BNoise is slightly higher than that of GANet. GANet+PNoise achieves much better AP score in terms of foggy scenes compared to GANet ($0.2118$ vs. $0.1509$), which demonstrates the effectiveness of the introduced Perlin noise. If we perform the full data augmentation strategy in our training samples, the overall performance will increase by $2\%$.

\subsubsection{Effectiveness of Background Attention.}
Different from the previous unsupervised attention modules, our Background Attention (BA) is guided based on discrimination between the background and objects. Firstly, we study different fusion strategies of the proposed BA in Figure~\ref{fig:fusion}, \ie, early fusion (EF), mixed fusion (MF) and late fusion (LF). The results presented in Table~\ref{tab:abl_BA} show the early fusion strategy (\ie, GANet+BPNoise+EF) achieves the best performance. Secondly, we also compare BA with several previous channel-wise attention modules including SE block~\cite{DBLP:conf/cvpr/HuSS18} and GC block~\cite{DBLP:journals/corr/abs-1904-11492}. For a fair comparison, we use the same early fusion strategy in Figure~\ref{fig:fusion}(a). Compared to the baseline FPN fusion strategy using lateral connection~\cite{DBLP:conf/cvpr/LinDGHHB17}, all the attention modules can improve the performance by learning the weights of different channels of feature maps. However, our BA module can learn additional discriminative information of background, resulting in the best AP score in the drone based scenes under different camera views (\ie, \textit{front-view}, \textit{side-view} and \textit{bird-view}).

\begin{table}[t]
\centering
\setlength{\tabcolsep}{2.5pt}
{
\caption{Influence of foreground attention.}
\label{tab:abl_FA}
\begin{tabular}{lccccc}
\hline
\toprule
Method & $\kappa$ &AP & AP$_\text{low}$ & AP$_\text{med}$ & AP$_\text{high}$   \\
\midrule
GANet+BPNoise    & -     &0.4181 &0.5565 &0.4867 & 0.1656 \\
\midrule
\multirow{5}*{GANet+BPNoise+FA}
 & 0         &0.4271      &0.5729     &0.4978  & 0.1698\\
 & 1         &\textbf{0.4411}      &\textbf{0.5900}     &\textbf{0.5146}  &\textbf{0.1751}\\
 & 2         &0.4391      &0.5869     &0.5130  & 0.1736\\
 & 3         &0.4372      &0.5817     &0.5128  & 0.1718\\
 & 4         &0.4347   &0.5764 &0.5116 & 0.1699\\
\bottomrule
\end{tabular}}
\end{table}
\subsubsection{Effectiveness of Foreground Attention.} We enumerate the threshold for Foreground Attention (FA) $\kappa$ in \eqref{equ:corner}, \ie, $\kappa=\{0,1,2,3,4\}$, to study its influence on the accuracy. As shown in Table~\ref{tab:abl_FA}, we can conclude that GANet with the FA module achieves the best AP score $0.4411$ when the threshold $\kappa=1$. If we remove FA, the detection performance will decrease to $0.4181$. It shows the effectiveness of the FA module.

\subsubsection{Variants of GANet.}
In Table~\ref{tab:variant}, we compare various variants of GANet that combine several components in the network. Using data argumentation strategy can improve the performance considerably in all the attributes. Either BA or FA can improve the performance by $3\sim4\%$. Moreover, the proposed method using both attentions and data argumentation strategy can boost the performance by approximate $8\%$ improvement in AP score compared to the baseline GANet method.

\section{Conclusion}
In the paper, we propose a novel guided attention network to deal with object detection and counting in drone based scenes. Specifically, we introduce both background and foreground attention modules to not only learn background discriminative representation but also consider local appearance of the object, resulting in better accuracy. The experiments on three challenging datasets demonstrate the effectiveness of our method. We plan to expand our method to multi-class object detection and counting for future work.

{\small
\bibliographystyle{ieee}
\bibliography{references}   
}

\end{document}